# Title: Deep Learning Microscopy

**Authors:** Yair Rivenson[1,2,3], Zoltán Göröcs[1,2,3]†, Harun Günaydın[1]†, Yibo Zhang[1,2,3], Hongda Wang[1,2,3], Aydogan Ozcan[1,2,3,4]*

**Affiliations:**

[1]Electrical Engineering Department, University of California, Los Angeles, CA, 90095, USA.

[2]Bioengineering Department, University of California, Los Angeles, CA, 90095, USA.

[3]California NanoSystems Institute (CNSI), University of California, Los Angeles, CA, 90095, USA.

[4]Department of Surgery, David Geffen School of Medicine, University of California, Los Angeles, CA, 90095, USA.

*Correspondence: ozcan@ucla.edu

Address: 420 Westwood Plaza, Engr. IV 68-119, UCLA, Los Angeles, CA 90095, USA

Tel: +1(310)825-0915

Fax: +1(310)206-4685

†Equal contributing authors.

**Abstract**:

We demonstrate that a deep neural network can significantly improve optical microscopy, enhancing its spatial resolution over a large field-of-view and depth-of-field. After its training, the only input to this network is an image acquired using a regular optical microscope, without any changes to its design. We blindly tested this deep learning approach using various tissue samples that are imaged with low-resolution and wide-field systems, where the network rapidly outputs an image with remarkably better resolution, matching the performance of higher numerical aperture lenses, also significantly surpassing their limited field-of-view and depth-of-field. These results are transformative for various fields that use microscopy tools, including e.g., life sciences, where optical microscopy is considered as one of the most widely used and deployed techniques. Beyond such applications, our presented approach is broadly applicable to other imaging modalities, also spanning different parts of the electromagnetic spectrum, and can be used to design computational imagers that get better and better as they continue to image specimen and establish new transformations among different modes of imaging.

## Introduction

Deep learning is a class of machine learning techniques that uses multi-layered artificial neural networks for automated analysis of signals or data [1,2]. The name comes from the general structure of deep neural networks, which consist of several layers of artificial neurons stacked over each other. One type of a deep neural network is the deep convolutional neural network (CNN). Typically, an individual layer of a deep convolutional network is composed of a convolutional layer and a non-linear operator. The kernels (filters) in these convolutional layers are randomly initialized and can then be trained to learn how to perform specific tasks using supervised or unsupervised machine learning techniques. CNNs form a rapidly growing research field with various applications in e.g., image classification[3], annotation[4], style transfer[5], compression[6], and deconvolution in photography[7–9], among others[10–13]. Recently, deep neural networks have also been used for optical phase recovery and holographic image reconstruction[14].

Here, we demonstrate the use of a deep neural network to significantly enhance the performance of an optical microscope without changing its design or hardware. This network uses a single image that is acquired under a standard microscope as input, and quickly outputs an improved image of the same specimen, e.g., in less than 1 sec using a laptop, matching the resolution of higher numerical aperture (NA) objectives, while at the same time surpassing their limited field-of-view (FOV) and depth-of-field (DOF). The first step in this deep learning based microscopy framework involves learning the statistical transformation between low-resolution and high resolution microscopic images, which is used to train a CNN. Normally, this transformation can be physically understood as a spatial convolution operation followed by an under-sampling step (going from a high resolution and high magnification microscopic image to a low-resolution and low magnification one). However, the proposed CNN framework is detached from the physics of light-matter interaction and image formation, and instead focuses on training of multiple layers of artificial neural networks to statistically relate low-resolution images (input) to high-resolution images (output) of a specimen. In fact, to train and blindly test this deep learning based imaging framework, we have chosen bright-field microscopy with spatially and temporally incoherent broadband illumination, which presents challenges to provide an exact analytical or numerical modelling of light-sample interaction and the related physical image formation process, making the relationship between high-resolution images and low-resolution ones significantly more complicated to exactly model or predict. Although bright-field microscopy images have been our focus in this manuscript, the same deep learning framework is broadly applicable to other microscopy modalities, including e.g., holography, dark-field, fluorescence, multi-photon, optical coherence tomography, among others.

## Results and Discussion

To initially train the deep neural network, we acquired microscopy images of Masson's trichrome stained lung tissue sections using a pathology slide, obtained from an anonymous pneumonia patient. The lower resolution images were acquired with a 40×/0.95NA objective lens providing a FOV of 150μm×150μm per image, while the higher resolution training images were acquired with a 100×/1.4NA oil-immersion objective lens providing a FOV of 60μm×60μm per image, i.e., 6.25-fold smaller in area. Both the low-resolution and high-resolution images were acquired with 0.55-NA condenser illumination leading to a diffraction limited resolution of ~0.36 μm and ~0.28μm, respectively, both of which were adequately sampled by the image sensor chip, with an 'effective' pixel size of ~0.18μm and ~0.07μm, respectively. Following a digital registration procedure to match the corresponding fields-of-view of each set of images (Supplementary Information), we generated 179 low-resolution images corresponding to different regions of the lung tissue sample, which were used as input to our network, together with their corresponding high-resolution labels for each FOV. Out of these images, 149 low-resolution input images and their corresponding high-resolution labels were randomly selected to be used as our training image set, while 10 low-resolution images and their corresponding high-resolution labels were used for selecting and validating the final network model, and the remaining 20 low-resolution inputs and their corresponding high-resolution labels formed our test images used to blindly quantify the average performance of the final network. This training dataset was further augmented by extracting 60×60-pixel and 150×150-pixel image patches with 40% overlap, from the low resolution and high resolution images, respectively, which effectively increased our training data size by more than 6-fold. As shown in Fig. 1a and further detailed in (Supplementary Information, Section 1), these training image patches were randomly assigned to 149 batches, each containing 64, randomly drawn, low and high-resolution image pairs, forming a total of 9,536 input patches for the network training process (see Supplementary Information, Section 5). The pixel count and the number of the image patches were empirically determined to allow rapid training of the network, while at the same time containing distinct sample features in each patch. In this training phase, as further detailed in Supplementary Information, we utilized an optimization algorithm to adjust the network's parameters using the training image set and utilized the validation image set to determine the best network model, also helping to avoid overfitting to the training image data.

After this training procedure, which needs to be performed only once, the CNN is fixed (Fig. 1b, and Supplementary Information, Section 1) and ready to blindly output high resolution images of samples of any type, i.e., not necessarily from the same tissue type that the CNN has been trained on. To demonstrate the success of this deep learning enhanced microscopy approach, first we blindly tested the network's model on entirely different sections of Masson's trichrome stained lung tissue, which were not used in our training process, and in fact were taken from another anonymous patient. These samples were imaged using the same 40×/0.95NA and 100×/1.4NA objective lenses with 0.55NA condenser illumination, generating various input images for our CNN. The output images of the CNN for these input images are summarized in Fig. 2, which clearly demonstrate the ability of the network to significantly enhance the spatial resolution of the input images, whether or not they were initially acquired with a 40×/0.95NA or a 100×/1.4NA objective lens. For the network output image shown in Fig. 2a, we used an input image acquired with a 40×/0.95NA objective lens and therefore it has a FOV that is 6.25-fold larger compared to the 100× objective FOV, which is highlighted with a red-box in Fig. 2a. Zoomed in regions of interest (ROI) corresponding to various input and output images are also shown in Fig. 2b-p better illustrating the fine spatial improvements in the network output images compared to the corresponding input images. To give an example on the computational load of this approach, the network output images shown in Fig. 2a and Fig. 2 c,h,m (with FOVs of 378.8 × 378.8 μm and 29.6 × 29.6 μm, respectively) took on average ~0.695 sec and 0.037 sec, respectively, to compute using a dual graphics processing unit (GPU) running on a laptop computer (Supplementary Information, Section 7).

In Fig. 2, we also illustrate that 'self-feeding' the output of the network as its new input significantly improves the resulting output image, as demonstrated in Fig. 2d,i,n. A minor disadvantage of this self-feeding approach is increased computation time, e.g., ~0.062 sec on average for Fig. 2d,i,n on the same laptop computer, in comparison to ~0.037 sec on average for Fig. 2c,h,m (Supplementary Information, Section 7). After one cycle of feeding the network with its own output, the next cycles of self-feeding do not change the output images in a noticeable manner, as also highlighted Supplementary Figure 5.

Quite interestingly, when we use the same deep neural network model on input images acquired with a 100×/1.4NA objective lens, the network output also demonstrates significant enhancement in spatial details that appear blurry in the original input images. These results are demonstrated in Fig. 2f,k,p and Supplementary Figure 6 revealing that the same learnt model (which was trained on the transformation of 40×/0.95NA images into 100×/1.4NA images) can also be used to super-resolve images that were captured with higher-magnification and higher numerical-aperture lenses compared to the input images of the training model. This feature suggests the scale-invariance of the image transformation (from lower resolution input images to higher resolution ones) that the CNN is trained on.

Next, we blindly applied the same lung tissue trained CNN for improving the microscopic images of a Masson's trichrome stained *kidney* tissue section obtained from an anonymous moderately advanced diabetic nephropathy patient. The network output images shown in Fig. 3 emphasize several important features of our deep learning based microscopy framework. First, this tissue type, although stained with the same dye (Masson's trichrome) is entirely new to our lung tissue trained CNN, and yet, the output images clearly show a similarly outstanding performance as in Fig. 2. Second, similar to the results shown in Fig. 2, self-feeding the output of the same lung tissue network as a fresh input back to the network further improves our reconstructed images, even for a kidney tissue that has not been part of our training process; see e.g., Fig. 3d,i,n. Third, the output images of our deep learning model also exhibit significantly larger DOF. To better illustrate this, the output image of the lung tissue trained CNN on a kidney tissue section imaged with a 40×/0.95NA objective was compared to an extended DOF image, which was obtained by using a depth-resolved stack of 5 images acquired using a 100×/1.4NA objective lens (with 0.4μm axial increments). To create the gold standard, i.e., the extended DOF image used for comparison to our network output, we merged these 5 depth-resolved images acquired with a 100×/1.4NA objective lens using a wavelet based depth-fusion algorithm[15]. The network's output images, shown in Fig. 3d,i,n, clearly demonstrate that several spatial features of the sample that appear in-focus in the deep learning output image can only be inferred by acquiring a depth-resolved stack of 100×/1.4NA objective images because of the shallow DOF of such high NA objective lenses – also see the yellow pointers in Fig. 3n and p to better visualize this DOF enhancement. Stated differently, the network output image not only has 6.25-fold larger FOV (~379 × 379 µm) compared to the images of a 100×/1.4NA objective lens, but it also exhibits a significantly enhanced DOF. The same extended DOF feature of the deep neural network image inference is further demonstrated using lung tissue samples shown in Fig. 2n and o.

Until now, we have focused on bright-field microscopic images of different tissue types, all stained with the same dye (Masson's trichrome), and used a deep neural network to blindly transform lower resolution images of these tissue samples into higher resolution ones, also showing significant enhancement in FOV and DOF of the output images. Next, we tested to see if a CNN that is trained on one type of stain can be applied to other tissue types that are stained with another dye. To investigate this, we trained a new CNN model (with the same network architecture) using microscopic images of a hematoxylin and eosin (H&E) stained human breast tissue section obtained from an anonymous breast cancer patient. As before, the training pairs were created from 40×/0.95NA lower resolution images and 100×/1.4NA high-resolution images (see Supplementary Tables 1,2 for specific implementation details). First, we blindly tested the results of this trained deep neural network on images of breast tissue samples (which were not part of the network training process) acquired using a 40×/0.95NA objective lens. Figure 4 illustrates the success of this blind testing phase, which is expected since this network has been trained on the same type of stain and tissue (i.e., H&E stained breast tissue). To compare, in the same Fig. 4 we also report the output images of a previously used deep neural network model (trained using lung tissue sections stained with the Masson's trichrome) for the same input images reported in Fig. 4. Except a relatively minor color distortion, all the spatial features of the H&E stained breast tissue sample have been resolved using a CNN trained on Masson's trichrome stained lung tissue. These results, together with the earlier ones discussed so far, clearly demonstrate the universality of the deep neural network approach, and how it can be used to output enhanced microscopic images of various types of samples, from different patients and organs and using different types of stains. A similarly outstanding result, with the same conclusion, is provided in Supplementary Figure 7, where the deep learning network trained on H&E stained breast tissue images was applied on Masson's trichrome stained lung tissue samples imaged using a 40×/0.95NA objective lens, representing the opposite case of Fig. 4.

Finally, to quantify the effect of our deep neural network on the spatial frequencies of the output image, we have applied the CNN that was trained using the lung tissue model on a resolution test target, which was imaged using a 100×/1.4NA objective lens, with a 0.55NA condenser. The objective lens was oil immersed as depicted in Supplementary Figure 8a, while the interface between the resolution test target and the sample cover glass was not oil immersed, leading to an effective NA of $\leq 1$ and a lateral diffraction limited resolution of $\geq 0.355$µm. The modulation transfer function (MTF) was evaluated by calculating the contrast of different elements of the resolution test target (Supplementary Information, Section 8). Based on this experimental analysis, the MTFs for the input image and the output image of the deep neural network that was trained on *lung tissue* are compared to each other in (Supplementary Information, Section 8). The output image of the deep neural network, despite the fact that it was trained on tissue samples imaged with a 40×/0.95NA objective lens, shows an increased modulation contrast for a significant portion of the spatial frequency spectrum, at especially high frequencies, while also resolving a period of 0.345 µm (Supplementary Information, Section 8).

To conclude, we have demonstrated how deep learning significantly enhances optical microscopy images, by improving their resolution, FOV and DOF. This deep learning approach is extremely fast to output an improved image, e.g., taking on average ~ 0.69 sec per image with a FOV of ~ 379 x 379 µm even using a laptop computer, and only needs a single image taken with a standard optical microscope without the need for extra hardware or user specified post-processing. After appropriate training, this framework is universally applicable to all forms of optical microscopy and imaging techniques and can be used to transfer images that are acquired under low resolution systems into high resolution and wide-field images, significantly extending the space bandwidth product of the output images. Furthermore, using the same deep learning approach we have also demonstrated the extension of the spatial frequency response of the imaging system along with an extended DOF. In addition to optical microscopy, this entire framework can also be applied to other computational imaging approaches, also spanning different parts of the electromagnetic spectrum, and can be used to design computational imagers with improved resolution, FOV and DOF.

**Methods**

*Sample Preparation*: De-identified formalin-fixed paraffin-embedded (FFPE) hematoxylin and eosin (H&E) stained human breast tissue section from a breast cancer patient, Masson's trichrome stained lung tissue section from 2 pneumonia patients, and Masson's trichrome stained kidney tissue section from a moderately advanced diabetic nephropathy patient were obtained from the Translational Pathology Core Laboratory at UCLA. Sample staining was done at the Histology Lab at UCLA. All the samples were obtained after de-identification of the patient and related information and were prepared from existing specimen. Therefore, this work did not interfere with standard practices of care or sample collection procedures.

*Microscopic Imaging*: Image data acquisition was performed using an Olympus IX83 microscope equipped with a motorized stage and controlled by MetaMorph® microscope automation software (Molecular Devices, LLC). The images were acquired using a set of Super Apochromat objectives, (UPLSAPO 40×2/0.95NA, 100×O/1.4NA – oil immersion objective lens). The color images were obtained using a Qimaging Retiga 4000R camera with a pixel size of 7.4 µm.

**Acknowledgements**

**Data and materials availability**: All the data and methods needed to evaluate the conclusions in this work are present in the main text and the Supplementary Information. Additional data related to this paper may be requested from the authors.

**Funding:** The Ozcan Research Group at UCLA acknowledges the support of the Presidential Early Career Award for Scientists and Engineers (PECASE), the Army Research Office (ARO; W911NF-13-1-0419 and W911NF-13-1-0197), the ARO Life Sciences Division, the National Science Foundation (NSF) CBET Division Biophotonics Program, the NSF Emerging Frontiers in Research and Innovation (EFRI) Award, the NSF EAGER Award, NSF INSPIRE Award, NSF Partnerships for Innovation: Building Innovation Capacity (PFI:BIC) Program, Office of Naval Research (ONR), the National Institutes of Health (NIH), the Howard Hughes Medical Institute (HHMI), Vodafone Americas Foundation, the Mary Kay Foundation, Steven & Alexandra Cohen Foundation, and KAUST. This work is based upon research performed in a laboratory renovated by the National Science Foundation under Grant No. 0963183, which is an award funded under the American Recovery and Reinvestment Act of 2009 (ARRA). Yair Rivenson is partially supported by the European Union's Horizon 2020 research and innovation programme under the Marie Skłodowska-Curie grant agreement No H2020-MSCA-IF-2014-659595 (MCMQCT).

**Author contributions**: Y.R. and A.O. conceived the research, Z.G., Y.R. and H.W. conducted the experiments, Y.R., H.G., Z.G. and Y.Z. processed the data. Y.R., H.G., Z.G. and A.O.

prepared the manuscript and all the other authors contributed to the manuscript. A.O. supervised the research.

**Competing financial interests**: None.

Correspondence and requests for materials should be addressed to Aydogan Ozcan, ozcan@ucla.edu

# Figures and Figure Captions

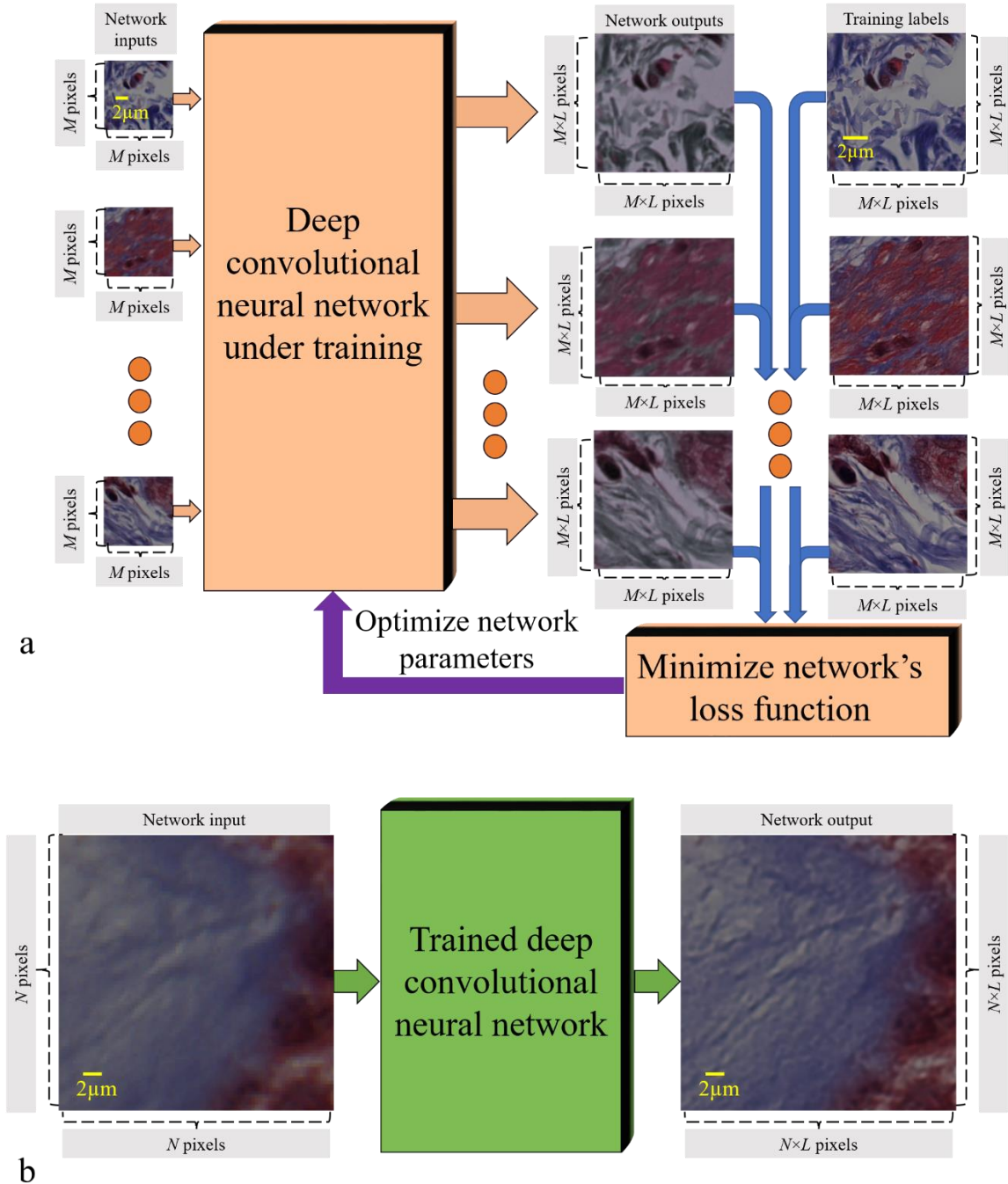

**Fig. 1**. Schematics of the deep neural network trained for microscopic imaging. **a**, The input is composed of a set of lower resolution images and the training labels are their corresponding high-resolution images. The deep neural network is trained by optimizing various parameters, which minimize the loss function between the network's output and the corresponding high-resolution training labels. **b**, After the training phase is complete, the network is blindly given an $N{\times}N$ pixel input image and rapidly outputs an $(N{\times}L){\times}(N{\times}L)$ image, showing improved spatial

resolution, field-of-view and depth-of-field.

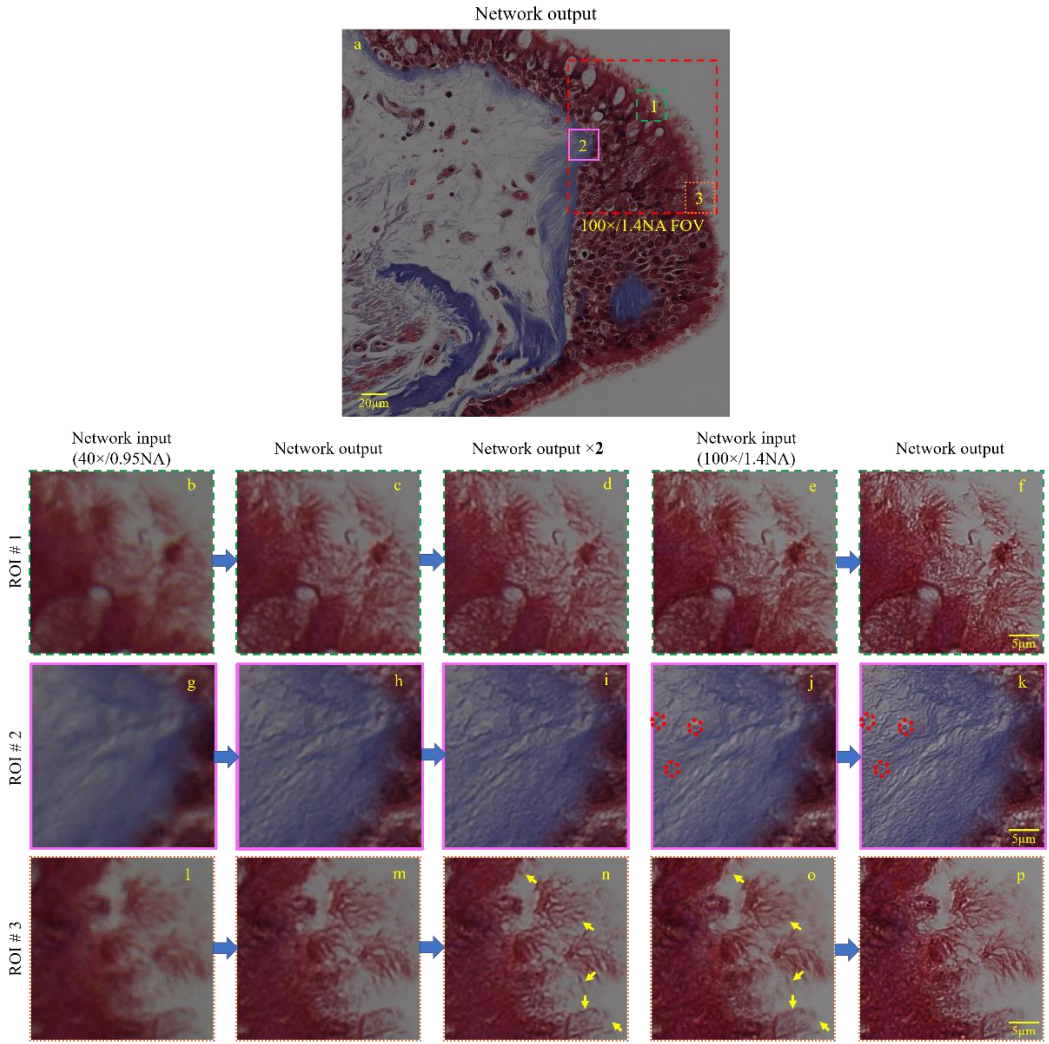

**Fig. 2**. Deep neural network output image corresponding to a Masson's trichrome stained lung tissue section taken from a pneumonia patient. The network was trained on images of a Masson's trichrome stained lung tissue sample taken from another patient. **a**, Image of the deep neural network output corresponding to a 40×/0.95NA input image. The red highlighted region denotes the FOV of a 100×/1.4NA objective lens. (**b**, **g**, **l**) Zoomed-in regions of interest (ROIs) of the input image (40×/0.95NA). (**c**, **h**, **m**) Zoomed-in ROIs of the neural network output image. (**d**, **i**, **n**) Zoomed-in ROIs of the neural network output image, taking the first output of the network, shown in **c**, **h** and **m**, as input. (**e**, **j**, **o**) Comparison images of the same ROIs, acquired using a 100×/1.4NA objective lens. (**f**, **k**, **p**) Result of the same deep neural network model applied on the 100×/1.4NA objective lens images (also see Fig. S6). The yellow arrows in **o** point to some of the out-of-focus features that are brought to focus in the network output image shown in N. Red circles in **j**, **k** point to some dust particles in the images acquired with our 100×/1.4NA

objective lens, and that is why they do not appear in **g-i**. The average network computation time for different ROIs is listed in Supplementary Table 3.

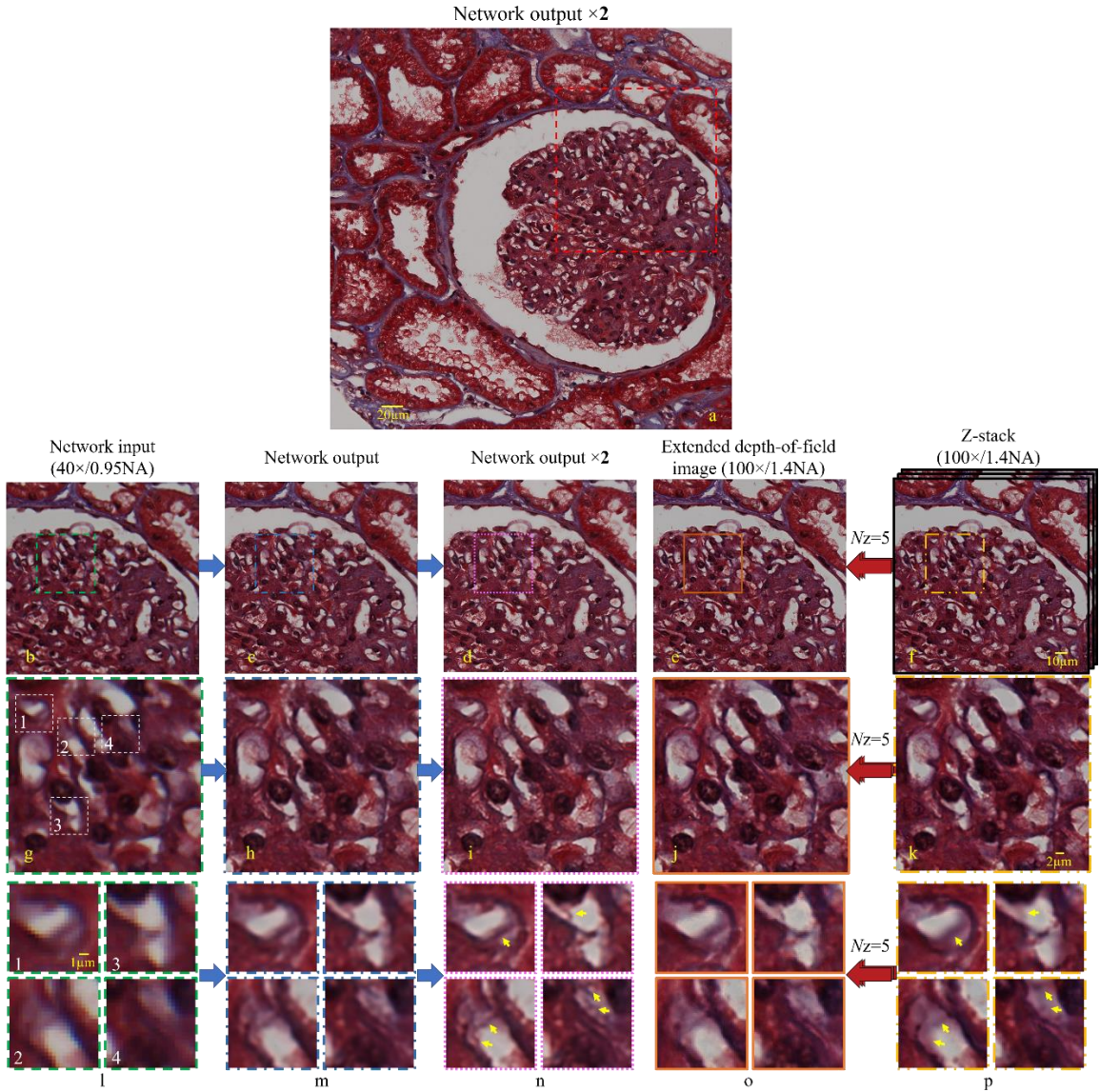

**Fig. 3**. Deep neural network output image of Masson's trichrome stained *kidney* tissue section obtained from a moderately advanced diabetic nephropathy patient. The network was trained on images of a Masson's trichrome stained *lung* tissue taken from another patient, **a**, Result of two-successive applications of the same deep neural network on a 40×/0.95NA image of the kidney tissue that is used as input. The red highlighted region denotes the FOV of a 100×/1.4NA objective lens. (**b**, **g**, **l**) Zoomed-in ROIs of the input image (40×/0.95NA). (**c**, **h**, **m**) Zoomed-in ROIs of the neural network output image, taking the corresponding 40×/0.95NA images as input. (**d**, **i**, **n**) Zoomed-in ROIs of the neural network output image, taking the first output of the network, shown in **c**, **h** and **m**, as input. (**e**, **j**, **o**) Extended depth-of-field image, algorithmically

calculated using $N_z = 5$ images taken at different depths using a 100×/1.4NA objective lens. (**f**, **k**, **p**) The auto-focused images of the same ROIs, acquired using a 100×/1.4NA objective lens. The yellow arrows in **p** point to some of the out-of-focus features that are brought to focus in the network output images shown in **n**. Also see Fig. S6.

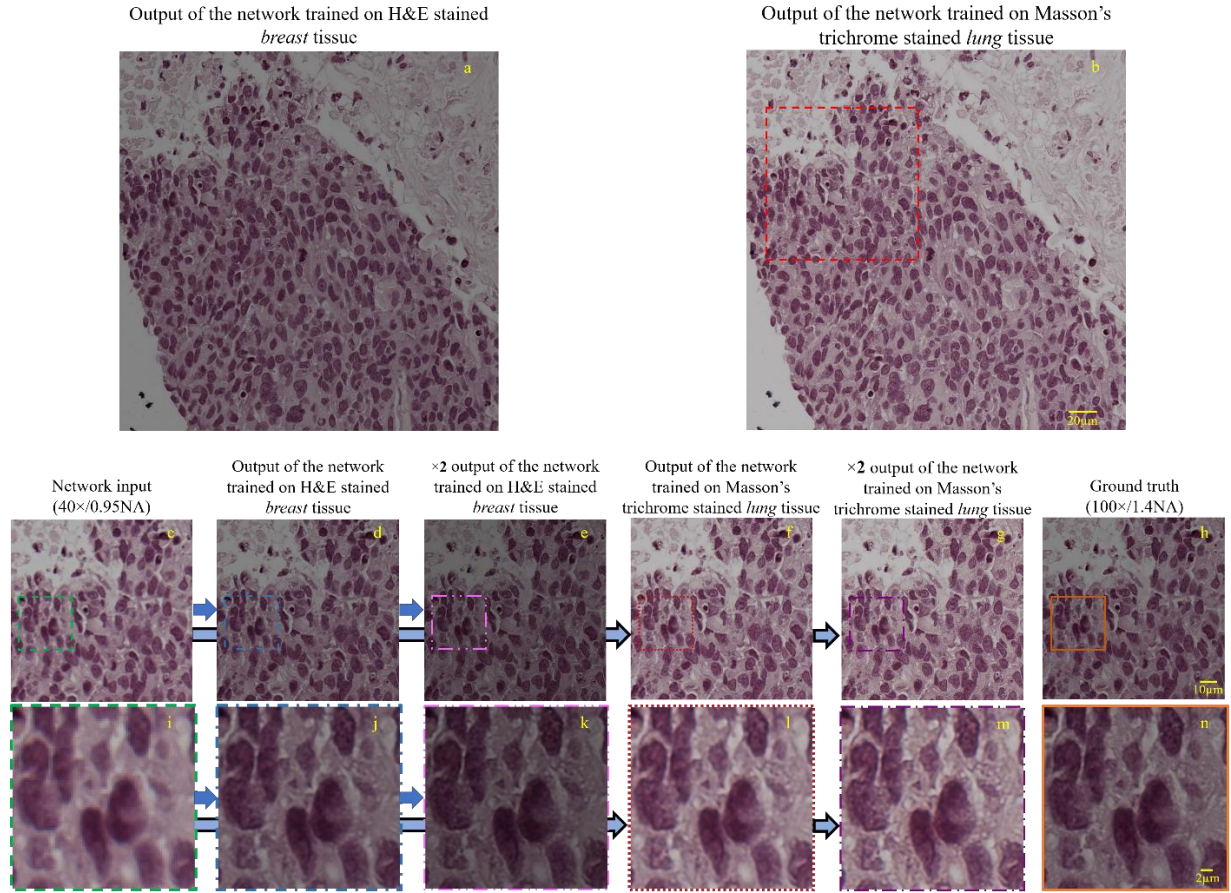

**Fig. 4.** Deep neural network based imaging of H&E stained breast tissue section. The output images of two different deep neural networks are compared to each other. The first network is trained on H&E stained breast tissue, taken from a different tissue section that is not used in the training phase. The second network is trained on a different tissue type and stain, i.e., Masson's trichrome stained lung tissue sections. (**c-n**) illustrate zoomed-in images of different ROIs of the input and output images, similar to Figs. 2-3. A similar comparison is also provided in Fig. S7.

# Supplementary Information - Deep Learning Microscopy

1. Deep Learning Network Architecture

The schematics of the architecture for training our deep neural network is depicted in Supplementary Fig. 1. The input images are mapped into 3 color channels: red, green and blue (RGB). The input convolutional layer maps the 3 input color channels, into 32 channels, as depicted in Supplementary Fig. 2. The number of output channels of the first convolutional layer was empirically determined to provide the optimal balance between the deep neural network's size (which affects the computational complexity and image output time) and its image transform performance. The input convolutional layer is followed by $K=5$ residual blocks[16]. Each residual block is composed of 2 convolutional layers and 2 rectified linear units (ReLU)[17,18], as shown in Supplementary Fig. 1. The ReLU is an activation function which performs $\text{ReLU}(x) = \max(0, x)$. The formula of each block can be summarized as:

$$X_{k+1} = X_k + \text{ReLU}(\text{ReLU}(X_k * W_k^{(1)}) * W_k^{(2)}), \tag{1}$$

where $*$ refers to convolution operation, $X_k$ is the input to the *k*-th block, $X_{k+1}$ denotes its output, $W_k^{(1)}$ and $W_k^{(2)}$ denote an ensemble of learnable convolution kernels of the *k*-th block, where the bias terms are omitted for simplicity. The output feature maps of the convolutional layers in the network are calculated as follows:

$$g_{k,j} = \sum_i f_{k,i} * w_{k,i,j} + \beta_{k,j} \Omega, \tag{2}$$

where $w_{k,i,j}$ is a learnable 2D kernel (i.e., the (*i*,*j*)-th kernel of $W_k$) applied to the *i*-th input feature map, $f_{k,i}$ (which is an *M*×*M*-pixel image in the residual blocks), $\beta_{k,j}$ is a learnable bias term, $\Omega$ is an *M*×*M* matrix with all its entries set as 1, and $g_{k,j}$ is the convolutional layer *j*-th output feature map (which is also an *M*×*M*-pixel image in the residual blocks). The size of all the kernels (filters) used throughout the network's convolutional layers is 3×3. To resolve the dimensionality mismatch of Eq. (2), prior to convolution, the feature map $f_{k,i}$ is zero-padded to a size of (*M*+2)×(*M*+2) pixels, where only the central *M*×*M*-pixel part is taken following the convolution with kernel $w_{k,i,j}$.

To allow high level feature inference we increase the number of features learnt in each layer, by gradually increasing the number of channels, using the pyramidal network concept[18]. Using such pyramidal networks helps to keep the network's width compact in comparison to designs that sustain a constant number of channels throughout the network. The channel increase formula was empirically set according to[19]:

$$A_k = A_{k-1} + \text{floor}((\alpha \times k) / K + 0.5) \tag{3}$$

where $A_0 = 32$, *k*=[1:5], which is the residual block number, *K*=5 is the total number of residual blocks used in our architecture and *α* is a constant that determines the number of channels that will be added at each residual block. In our implementation, we used *α*=10, which yields $A_5 = 62$ channels at the output of the final residual block. In addition, we utilized the concept of residual connections (shortcutting the block's input to its output, see Supplementary Fig. 1), which was demonstrated to improve the training of deep neural networks by providing a clear path for information flow[18] and speed up the convergence of the training phase. Nevertheless, increasing the number of channels at the output of each layer leads to a dimensional mismatch between the inputs and outputs of a block, which are element-wise summed up in Eq. (1). This dimensional mismatch is resolved by augmenting each block's input channels with zero valued channels, which virtually equalizes the number of channels between a residual block input and output.

In our experiments, we have trained the deep neural network to extend the output image space-bandwidth-product by a non-integer factor of $L^2=2.5^2=6.25$ compared to the input images. To do so, first the network learns to enhance the input image by a factor of 5×5 pixels followed by a learnable down-sampling operator of 2×2, to obtain the desired $L$=2.5 factor (see Supplementary Fig. 3). More specifically, at the output of the $K$-th residual block $A_K = A_5 = 62$ channels are mapped to $3\times 5^2 = 75$ channels (Supplementary Fig. 3), followed by resampling of these 75 $(M \times M)$ pixels channels to three channels with $(M \times 5) \times (M \times 5)$ pixels grid[13,20]. These three $(M \times 5) \times (M \times 5)$ pixels channels are then used as input to an additional convolutional layer (with learnable kernels and biases, as the rest of the network), that two-times down-samples these images to three $(M \times 2.5) \times (M \times 2.5)$ color pixels. This is performed by using a two-pixel stride convolution, instead of a single pixel stride convolution, as performed throughout the other convolutional layers of the network. This way, the network learns the optimal down-sampling procedure for our microscopic imaging task. It is important to note that during the testing phase, if the number of input pixels to the network is odd, the resulting number of output image pixels will be determined by the ceiling operator. For instance, a 555×333-pixel input image will result in a 1388×833-pixel image for $L$=2.5.

The above-discussed deep network architecture provides two major benefits: first, the up-sampling procedure becomes a learnable operation with supervised learning, and second, using low resolution images throughout the network's layers makes the time and memory complexities of the algorithm $L^2$ times smaller[13] when compared to approaches that up-sample the input image as a precursor to the deep neural network. This has a positive impact on the convergence speed of both the training and image transformation phases of our network.

2. Data Pre-processing

To achieve optimal results, the network should be trained with accurately aligned low-resolution input images and high-resolution label image data. We match the corresponding input and label image pairs using the following steps: (A) Color images are converted to grayscale images. (B) A large field-of-view image is formed by stitching a set of low resolution images. (C) Each high-resolution label image is down-sampled (bicubic) by a factor $L$. This down-sampled image is used as a template image to find the highest correlation matching patch in the low-resolution stitched image. The highest correlating patch from the low-resolution stitched image is then digitally cropped. This cropped low-resolution image and the original high-resolution image, form an input-label pair, which is used for the network's training and testing. (D) Additional alignment is then performed on each of the input-label pairs to further refine the input-label matching, mitigating rotation, translation and scaling discrepancies between the lower resolution and higher resolution images.

3. Network Training

The network was trained by optimizing the following loss function ($\ell$) given the high-resolution training labels $Y^{HR}$:

$$\ell(\Theta) = \frac{1}{3 \times M^2 \times L^2} \sum_{c=1}^{3} \sum_{u=1}^{M \times L} \sum_{v=1}^{M \times L} \left\| Y_{c,u,v}^{\Theta} - Y_{c,u,v}^{HR} \right\|^2 + \lambda \frac{1}{3 \times M^2 \times L^2} \sum_{c=1}^{3} \sum_{u=1}^{M \times L} \sum_{v=1}^{M \times L} \left| \nabla Y^{\Theta} \right|_{c,u,v}^{2}, \quad (4)$$

where $Y_{c,u,v}^{\Theta}$ and $Y_{c,u,v}^{HR}$ denote the $u,v$-th pixel of the $c$-th color channel (where in our implementation we use three color channels, RGB) of the network's output image and the high resolution training label image, respectively. The network's output is given by $Y^{\Theta} = F(X_{input}^{LR}; \Theta)$, where $F$ is the deep neural network's operator on the low-resolution input image $X_{input}^{LR}$ and $\Theta$ is the network's parameter space (e.g., kernels, biases, weights). Also, $(M \times L) \times (M \times L)$ is the total number of pixels in each color channel, $\lambda$ is a regularization parameter, empirically set to ~0.001. $\left| \nabla Y^{\Theta} \right|_{c,u,v}^{2}$ is $u,v$-th pixel of the $c$-th color channel of the network's output image gradient[21], applied separately for each color channel, which is defined as: $\left| \nabla Y^{\Theta} \right|^2 = \left( h * Y^{\Theta} \right)^2 + \left( h^T * Y^{\Theta} \right)^2$, with:

$$h = \begin{bmatrix} -1 & 0 & 1 \\ -2 & 0 & 2 \\ -1 & 0 & 1 \end{bmatrix}, \quad (5)$$

and $(.)^T$ refers to the matrix transpose operator.

The above defined loss function balances between the mean-squared-error (MSE) and the image sharpness with a regularization parameter, $\lambda$. The MSE is used as a data fidelity term and the $l_2$-norm image gradient approximation helps mitigating the spurious edges that result from the pixel up-sampling process. Following the estimation of the loss function, the error is backpropagated through the network, and the network's parameters are learnt by using the Adaptive Moment Estimation (ADAM) optimization[22], which is a stochastic optimization method, that we empirically set a learning rate parameter of $10^{-4}$ and a mini-batch size of 64 image patches (Supplementary Table 2). All the kernels (for instance $w_{k,i,j}$) used in convolutional layers have 3×3 elements and their entries are initialized using truncated normal distribution with 0.05 standard deviation and 0 mean[16]. All the bias terms (for instance, $\beta_{k,j}$) are initialized with 0.

4. Network Testing

A fixed network architecture, following the training phase is shown in Supplementary Fig. 4, which receives an input of $P \times Q$-pixel image and outputs a $\lceil (P \times L) \rceil \times \lceil (Q \times L) \rceil$-pixel image, where $\lceil . \rceil$ is the ceiling operator. To numerically quantify the performance of our trained network models, we independently tested it using validation images, as detailed in Supplementary Table 2. The output images of the network were quantified using the structural similarity index[23] (SSIM). SSIM, which has a scale between 0 and 1, quantifies a human observer's perceptual loss from a gold standard image by taking into account the relationship among the contrast, luminance, and structure components of the image. SSIM is defined as 1 for an image that is identical to the gold standard image.

5. Implementation Details

The program was implemented using Python version 3.5.2, and the deep neural network was implemented using TensorFlow framework version 0.12.1 (Google). We used a laptop computer with Core i7-6700K CPU @ 4GHz (Intel) and 64GB of RAM, running a Windows 10 professional operating system (Microsoft). The network training and testing were performed using GeForce GTX 1080 GPUs (NVidia). For the training phase, using a dual-GPU configuration resulted in ~33% speedup compared to training the network with a single GPU. The training time of the deep neural networks for the lung and breast tissue image datasets is summarized in Table Supplementary Table 2 (for the dual-GPU configuration).

Following the conclusion of the training stage, the fixed deep neural network intakes an input stream of 100 low-resolution images each with 2,048×2,048-pixels, and outputs for each input image a 5,120×5,120-pixel high-resolution image at a total time of ~119.3 seconds (for all the 100 images) on a single laptop GPU. This runtime was calculated as the average of 5 different runs. Therefore, for $L$=2.5 the network takes 1.193 sec per output image on a single GPU. When employing a dual-GPU for the same task, the average runtime reduces to 0.695 sec per 2,048×2,048-pixel input image (see Supplementary Table 3 for additional details on the network output runtime corresponding to other input image sizes, including self-feeding of the network output).

6. Modulation Transfer Function (MTF) Analysis

To quantify the effect of our deep neural network on the spatial frequencies of the output image, we have applied the CNN that was trained using the Masson's trichrome stained lung tissue samples on a resolution test target (Extreme USAF Resolution Target on 4×1 mm Quartz Circle Model 2012B, Ready Optics), which was imaged using a 100×/1.4NA objective lens, with a 0.55NA condenser. The objective lens was oil immersed as depicted in Supplementary Fig. 8a, while the interface between the resolution test target and the sample cover glass was not oil immersed, leading to an effective objective NA of ≤1 and a lateral diffraction limited resolution ≥ 0.354µm (assuming an average illumination wavelength of 550 nm). MTF was evaluated by calculating the contrast of different elements of the resolution test target[24]. For each element, we horizontally averaged the resulting image along the element lines (~80-90% of the line length). We then located the center pixels of the element's minima and maxima and used their values for contrast calculation. To do that, we calculated the length of the element's cross-section from the resolution test target group and element number in micrometers, cut out a corresponding cross section length from the center of the horizontally averaged element lines. This also yielded the center pixel locations of the element's local maximum values (2 values) and minimum values (3 values). The maximum value, $I_{max}$, was set as the maximum of the local maximum values and the minimum value, $I_{min}$, was set as the minimum of the local minimum values. For the elements, where the minima and maxima of the pattern matched their calculated locations in the averaged cross section, the contrast value was calculated as: $(I_{max} - I_{min})/(I_{max} + I_{min})$. For the elements where the minima and maxima were not at their expected positons, thus the modulation of the element was not preserved, we set the contrast to 0. Based on this experimental analysis, the calculated contrast values are given Supplementary Table 4 and the MTFs for the input image and the output image of the deep neural network (trained on Masson's trichrome lung tissue) are compared to each other in Supplementary Fig. 8e.

**Supplementary Figures**

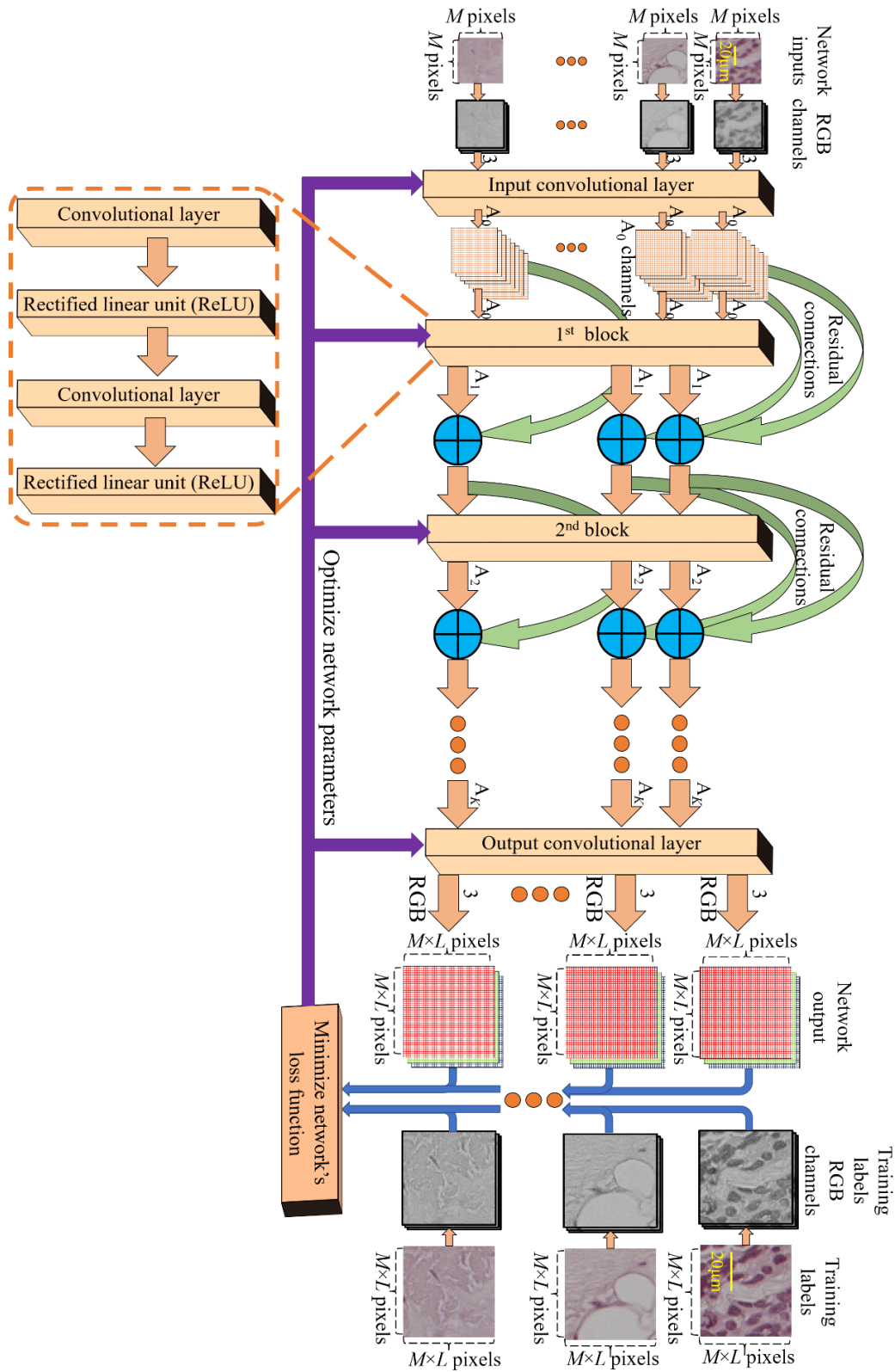

**Supplementary Figure 1.** Detailed schematics of the deep neural network training phase.

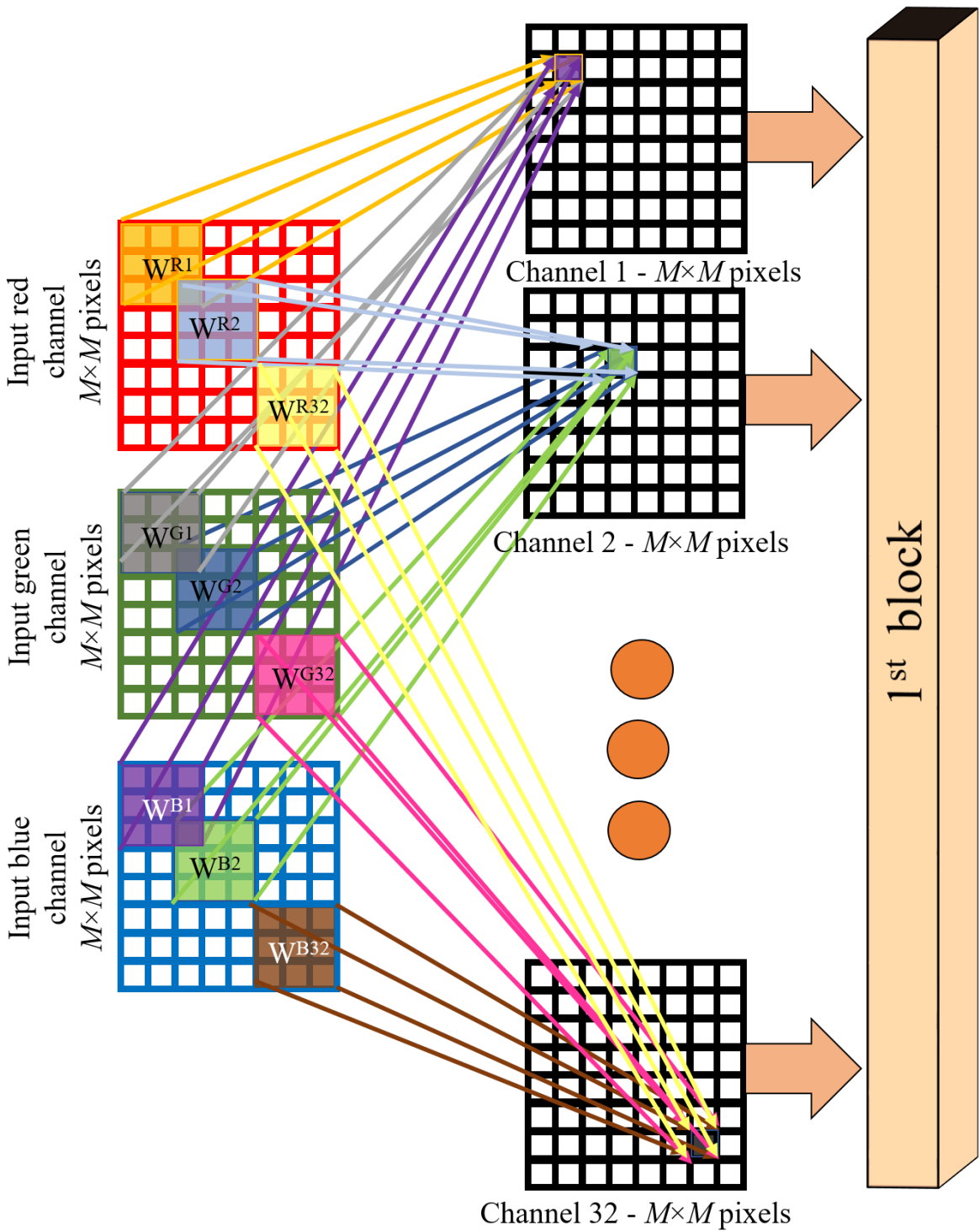

**Supplementary Figure 2.** Detailed schematics of the input layer of the deep neural network.

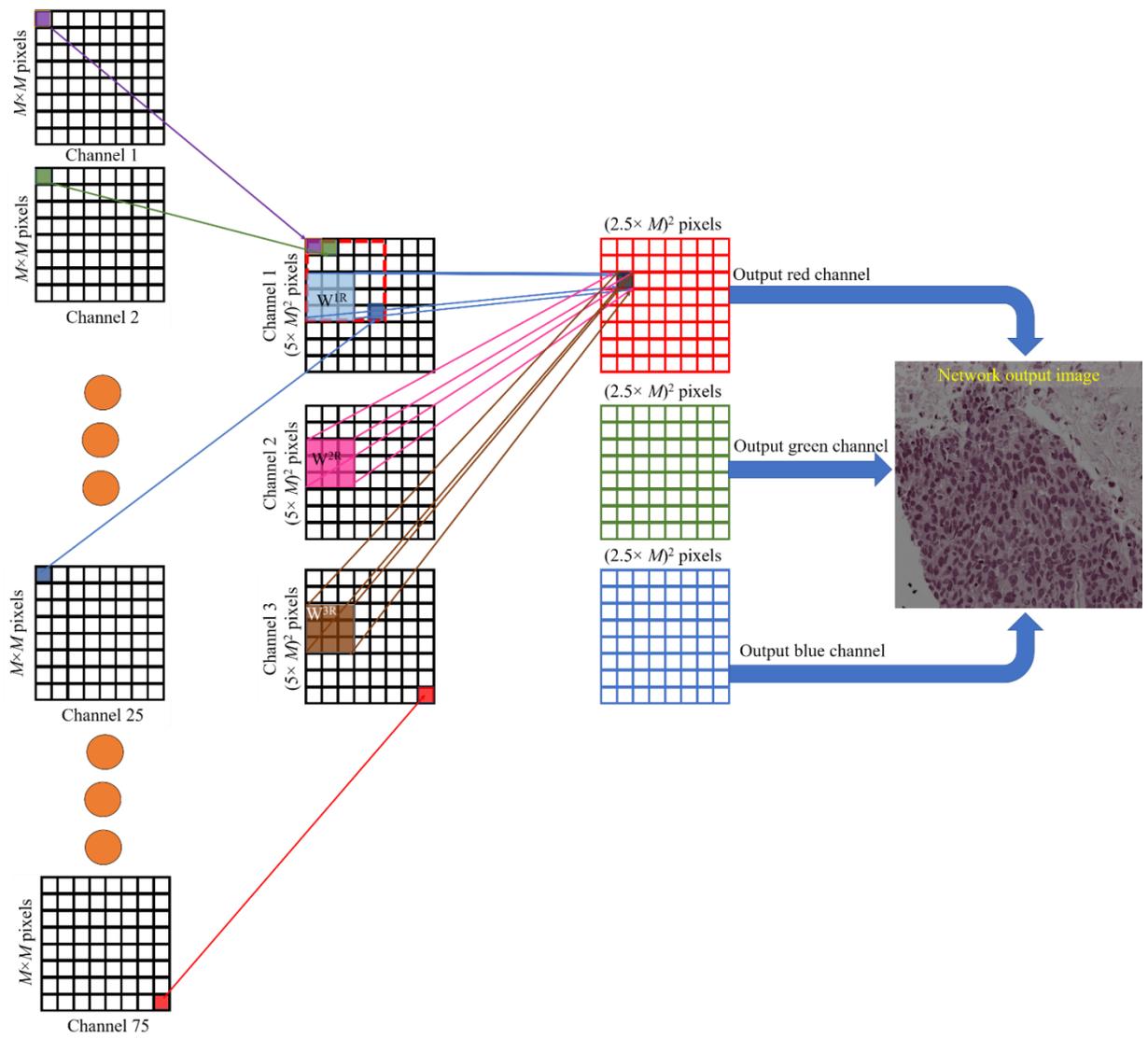

**Supplementary Figure 3.** Detailed schematics of the output layer of the deep neural network for *L*=2.5.

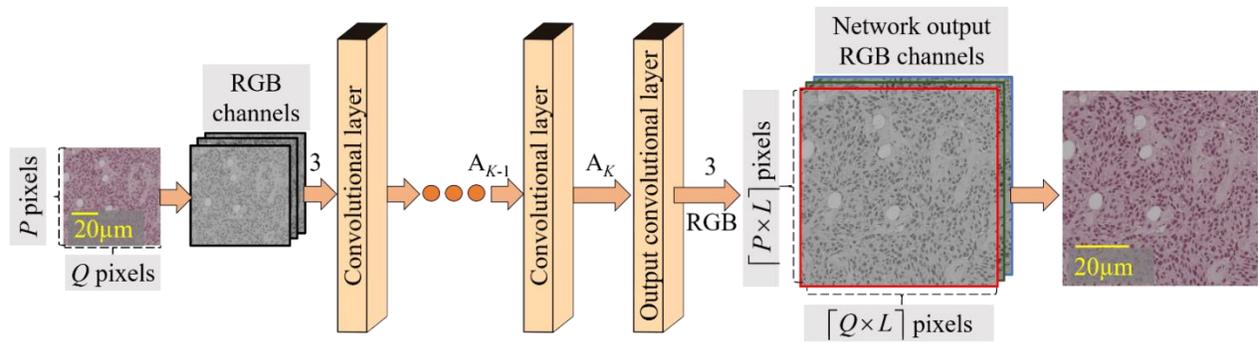

**Supplementary Figure 4.** Detailed schematics of the deep neural network high-resolution image inference (i.e., the testing phase).

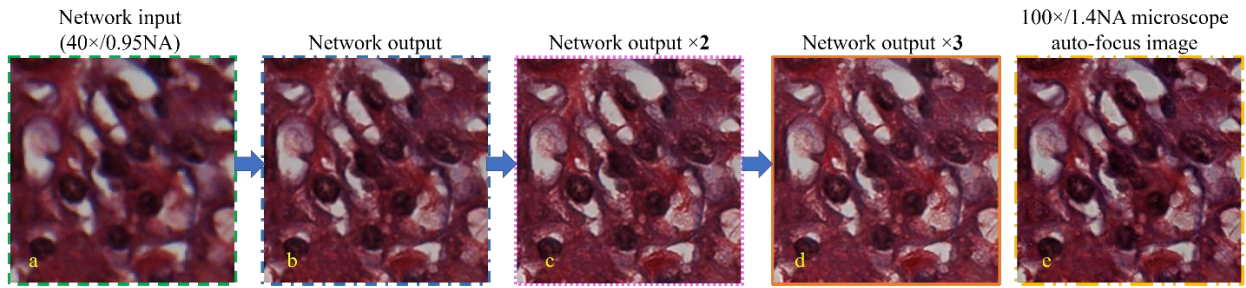

**Supplementary Figure 5.** Result of applying the deep neural network in a cyclic manner on Masson's trichrome stained kidney section images. **a**, Input image acquired with a 40×/0.95NA objective lens. The deep neural network is applied on this input image once, twice and three times, where the results are shown in **b**, **c** and **d**, respectively. **e**, 100×/1.4NA image of the same field-of-view is shown for comparison.

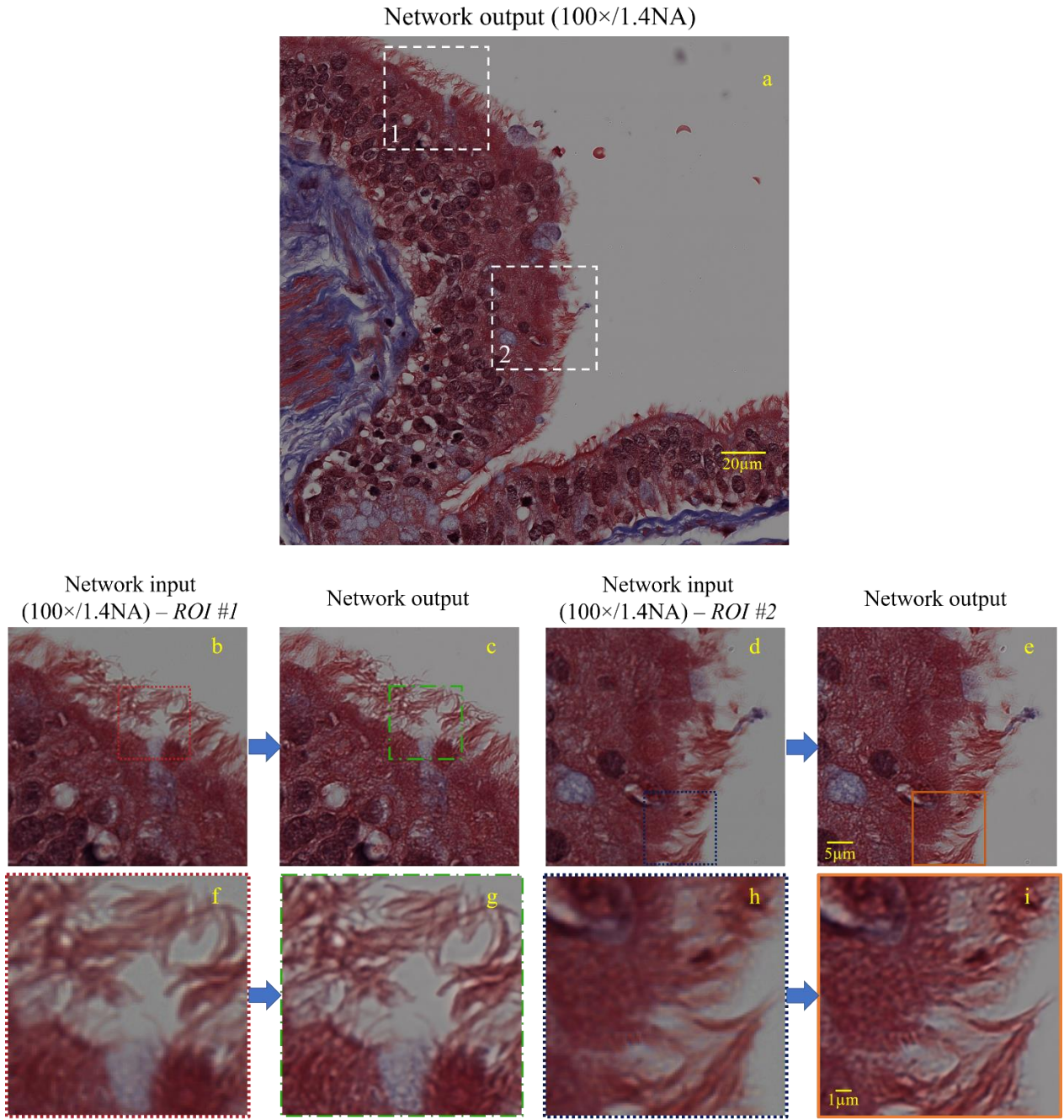

**Supplementary Figure 6.** Deep neural network output image corresponding to a Masson's trichrome stained lung tissue section taken from a pneumonia patient. The network was trained on images of a Masson's trichrome stained lung tissue taken from a different tissue block that was not used as part of the CNN training phase. **a**, Image of the deep neural network output corresponding to a 100×/1.4NA input image. (**b**, **f**, **d**, **h**) Zoomed-in ROIs of the input image (100×/1.4NA). (**c**, **g**, **e**, **i**) Zoomed-in ROIs of the neural network output image.

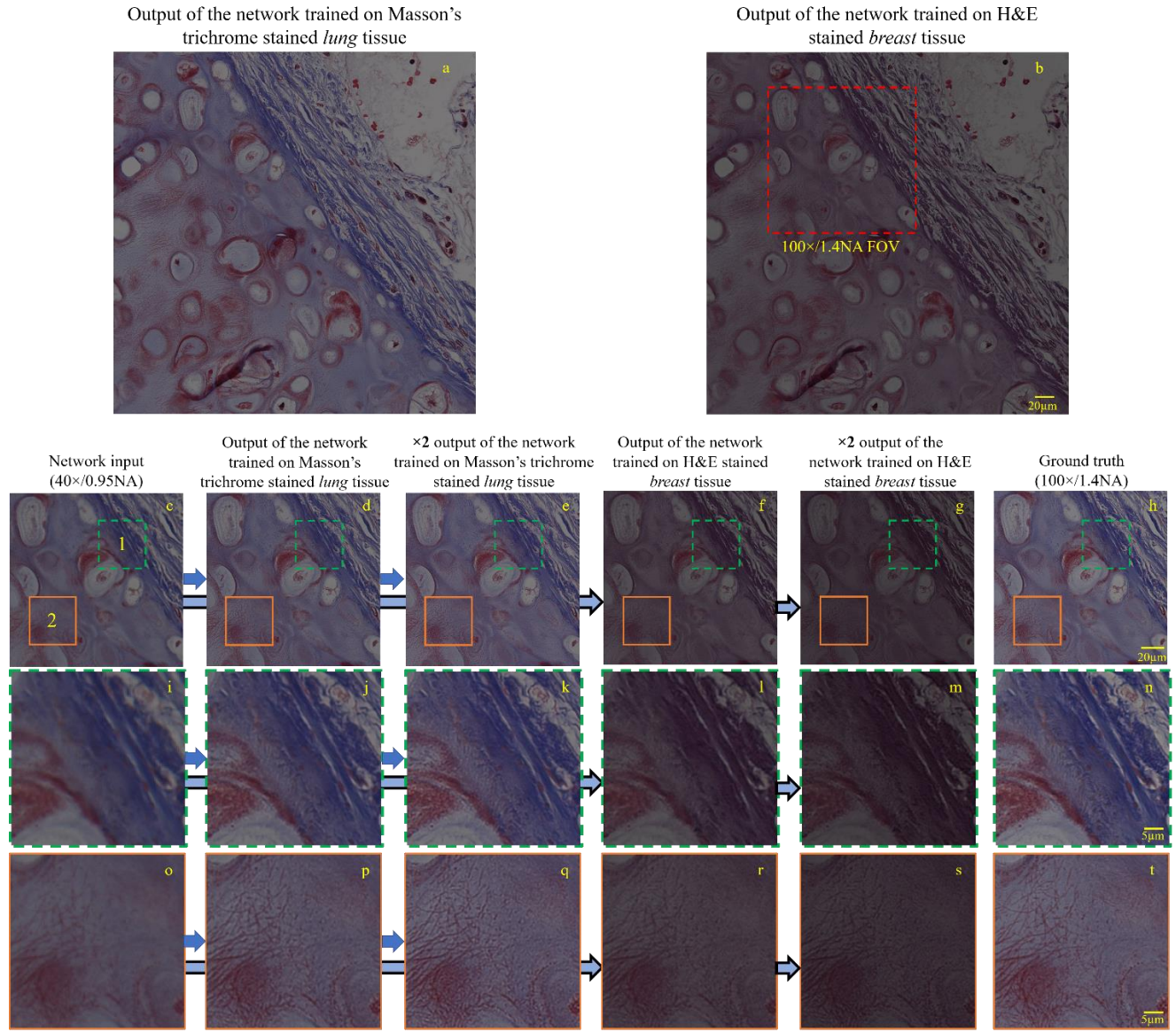

**Supplementary Figure 7. a**, Result of applying the *lung* tissue trained deep neural network model on a 40×/0.95NA *lung* tissue input image. **b**, Result of applying the *breast* tissue trained deep neural network model on a 40×/0.95NA *lung* tissue input image. (**c**, **i**, **o**) Zoomed in ROIs corresponding to the 40×/0.95NA input image. (**d**, **j**, **p**) Neural network output images, corresponding to input images **c**, **i** and **o**, respectively; the network is trained with lung tissue images. (**e**, **k**, **q**) Neural network output images, corresponding to input images **d**, **j**, and **p**, respectively; the network is trained with lung tissue images. (**f**, **l**, **r**) Neural network output images, corresponding to input images **c**, **i** and **o**, respectively; the network is trained with breast tissue images stained with a different dye, H&E. (**g**, **m**, **s**) Neural network output images, corresponding to input images **f**, **l**, and **r**, respectively; the network is trained with breast tissue images stained with H&E. (**h**, **n**, **t**) Comparison images of the same ROIs acquired using a 100×/1.4NA objective lens.

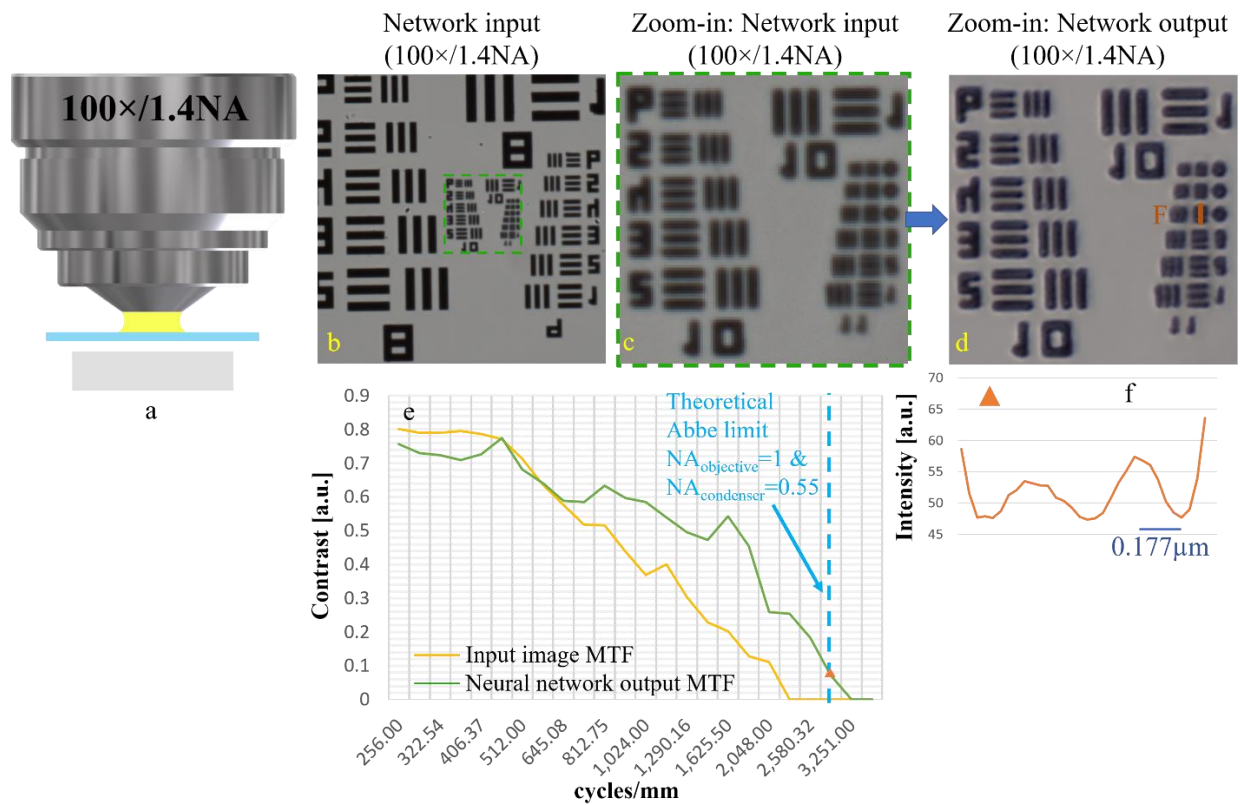

**Supplementary Figure 8**. Modulation transfer function (MTF) comparison for the input image and the output image of a deep neural network that is trained on images of a lung tissue section. **a**, Experimental apparatus: the US air-force (USAF) resolution target lies on a glass slide, with an air-gap in-between, leading to an effective numerical aperture of ≤ 1. The resolution test target was illuminated using a condenser with a numerical aperture of 0.55, leading to lateral diffraction limited resolution of ≥ 0.355µm. **b**, Input image acquired with a 100×/1.4NA. **c,** Zoom-in on the green highlighted region of interest highlighted in (**b**). **d**, Output image of the deep neural network applied on (**b, c**). **e**, MTF calculated from the input and output images of the deep network. **f**, Cross-sectional profile of group 11, element 4 (period: 0.345µm) extracted from the network output image shown in (**d**).

**Supplementary Tables**

|  | Test set | Bicubic up-sampling SSIM | Deep neural network SSIM |
|---|---|---|---|
| Masson's trichrome stained lung tissue | 20 images (224×224 pixels) | 0.672 | 0.796 |
| H&E stained breast tissue | 7 images (660×660 pixels) | 0.685 | 0.806 |

**Supplementary Table 1.** Average structural similarity index (SSIM) for the Masson's trichrome stained lung tissue and H&E stained breast tissue datasets, comparing bicubic up-sampling and the deep neural network output.

|  | Number of input-output patches (number of pixels for each low-resolution image) | Validation set (number of pixels for each low-resolution image) | Number of epochs till convergence | Training time |
|---|---|---|---|---|
| Masson's trichrome stained lung tissue | 9,536 patches (60×60 pixels) | 10 images (224×224 pixels) | 630 | 4hr, 35min |
| H&E stained breast tissue | 51,008 patches (60×60 pixels) | 10 images (660×660 pixels) | 460 | 14hr, 30min |

**Supplementary Table 2.** Deep neural network training details for the Masson's trichrome stained lung tissue and H&E stained breast tissue datasets.

|  |  | Single GPU runtime (*sec*) | | Dual GPU runtime (*sec*) | |
|---|---|---|---|---|---|
| Image FOV | Number of Pixels (input) | Network Output | Network Output x2 (Self-feeding) | Network Output | Network Output x2 (Self-feeding) |
| 378.8 × 378.8 µm (e.g., Fig. 2A) | 2048×2048 | 1.193 | 8.343 | 0.695 | 4.615 |
| 151.3 × 151.3 µm (e.g., red box in Fig. 2A) | 818×818 | 0.209 | 1.281 | 0.135 | 0.730 |
| 29.6 × 29.6 µm (e.g., Figs. 2B-L) | 160×160 | 0.038 | 0.081 | 0.037 | 0.062 |

**Supplementary Table 3.** Average runtime for different regions-of-interest shown in Fig. 2.

| Period (Cycles/mm) | 100×/1.4NA input contrast (a.u.) | Network output contrast (a.u.) |
|---|---|---|
| 256 | 0.801 | 0.756 |
| 287.350 | 0.790 | 0.729 |
| 322.539 | 0.790 | 0.724 |
| 362.038 | 0.795 | 0.709 |
| 406.374 | 0.787 | 0.726 |
| 456.140 | 0.771 | 0.774 |
| 512 | 0.713 | 0.681 |
| 574.700 | 0.636 | 0.640 |
| 645.079 | 0.577 | 0.588 |
| 724.077 | 0.517 | 0.585 |
| 812.749 | 0.516 | 0.634 |
| 912.280 | 0.439 | 0.597 |
| 1024 | 0.369 | 0.585 |
| 1290.159 | 0.303 | 0.538 |
| 1448.154 | 0.229 | 0.473 |
| 1625.498 | 0.201 | 0.542 |
| 1824.560 | 0.128 | 0.455 |
| 2048 | 0.111 | 0.259 |
| 2298.802 | 0 | 0.254 |
| 2580.318 | 0 | 0.1827 |
| 2896.309 | 0 | 0.072 |
| 3250.997 | 0 | 0 |
| 3649.121 | 0 | 0 |

**Supplementary Table 4.** Calculated contrast values for the USAF resolution test target elements.